\documentclass[10pt]{article}

\usepackage[margin=0.75in]{geometry}
\usepackage{microtype}
\usepackage{amsmath,amssymb,amsfonts}
\usepackage{booktabs}
\usepackage{array}
\usepackage{tabularx}
\usepackage{graphicx}
\usepackage{xcolor}
\usepackage{hyperref}
\usepackage[nameinlink,noabbrev]{cleveref}
\usepackage[numbers,sort&compress]{natbib}
\usepackage{morefloats}
\usepackage[section]{placeins}

\graphicspath{{figures/}}

\hypersetup{
  colorlinks=true,
  linkcolor=blue!50!black,
  citecolor=blue!50!black,
  urlcolor=blue!50!black
}

\newcommand{\routeacc}{\texttt{route\_acc}}
\newcommand{\tokacc}{\texttt{gen\_tok\_acc\_full}}
\newcommand{\exactacc}{\texttt{gen\_exact\_full}}
\newcolumntype{L}[1]{>{\raggedright\arraybackslash}p{#1}}
\newcolumntype{Y}{>{\raggedright\arraybackslash}X}
\title{Fixed-Interface State Transfer Reveals Reuse in Controlled Routing}
\author{
\parbox[t]{0.42\textwidth}{\centering
Yanzhen Lu\\[0.4ex]
\small Shanghai Qizhi Institute\\[0.2ex]
\small \texttt{luyanzhen@sqz.ac.cn}
}\hfill
\parbox[t]{0.42\textwidth}{\centering
Zhicheng Qian\\[0.4ex]
\small Southeast University\\[0.2ex]
\small \texttt{220232282@seu.edu.cn}
}\\[6.0ex]
\parbox[t]{0.42\textwidth}{\centering
Muchen Jiang\\[0.4ex]
\small Nanjing Whale Cloud\\[0.2ex]
\small \texttt{jiang.muchen2@iwhalecloud.com}
}\hfill
\parbox[t]{0.42\textwidth}{\centering
Xingyu Zhou\\[0.4ex]
\small Xiamen University\\[0.2ex]
\small \texttt{33520241153320@stu.xmu.edu.cn}
}
}
\date{}

\begin{document}
\maketitle

\begin{abstract}
Prompt-based interventions can change model behavior, but trained success alone does not identify where the behaviorally relevant state is represented.
We study this question in controlled routing tasks using interfaces chosen on support data, held-out query evaluation, and matched necessity, sufficiency, and wrong-interface controls.
On GPT-2 triop, an early interface supports exact transfer under these tests.
On GPT-2 add/sub, zero-retrain compiled transfer at the fixed interface recovers most of donor routing accuracy, while trainable prompt slots can relearn the same behavior at several other positions only after additional support examples and optimization.
These results distinguish fixed-interface reuse from prompt relocation in a setting where the two can be tested directly.
Qwen routing provides a cross-architecture consistency check for the same matched-interface pattern at the operator token, although donor-specific identity on the local \(V\)-path remains unresolved.
Generation and reasoning branches are used to map scope: they show broader transport or weaker controller identifiability once control depends on longer trajectories or harder selection.
In controlled routing, fixed-interface transfer is therefore stronger evidence of reuse than trained prompt success alone.
\end{abstract}

\section{Introduction}
Prompt tuning, prefix tuning, and representation steering can all change language-model behavior \citep{lester2021prompttuning,li2021prefixtuning,hu2021lora,turner2023actadd,zou2023repe}, but trained success alone does not show where the endogenous control variable lives.
A learned prompt can create a useful internal signal, relocate one, or reuse one that is already available at runtime.
The central question of this paper is whether prompt-induced routing behavior can be traced to reuse of a state at a fixed internal interface before any retraining takes place.

When a state chosen on support data can be transferred at one fixed interface and survives held-out necessity, sufficiency, and wrong-interface tests, it provides stronger causal evidence for reuse than any trained prompt that is allowed to learn a new state elsewhere.
The strongest evidence appears in controlled routing: GPT-2 triop gives exact transfer at a single early interface, GPT-2 add/sub shows that zero-retrain transfer and learned prompt relocation are functionally distinct, and Qwen routing provides a cross-architecture consistency check for the same matched-interface pattern at the operator token.
Generation and reasoning branches then delineate scope by showing how the picture changes once control depends on broader trajectory state or harder selection.

We first define the evidence standard and locking protocol, then present the routing results, then examine generation and reasoning extensions, and finally collect unresolved issues in the limitations.

\paragraph{A simple example.}
The core comparison is easiest to see in a small routing setting.
Suppose a \emph{donor} prompt produces the correct route and a \emph{receiver} prompt does not.
If copying one internal state from donor to receiver at one fixed interface makes the receiver route correctly on held-out queries, and the reverse replacement removes that behavior, then the evidence supports reuse of a pre-existing state at that interface.
If the same endpoint can instead be recovered only after training a new prompt slot, then the result shows that the behavior can be rebuilt, but not that the original interface was already carrying the relevant state.
The routing sections instantiate this comparison directly.

\section{Related Work}
Prompt and prefix methods establish that compact control signals can steer model behavior \citep{lester2021prompttuning,li2021prefixtuning,hu2021lora},
while activation steering and related intervention methods show that latent directions can change model behavior without full finetuning \citep{turner2023actadd,zou2023repe}.
Model editing and context-sensitivity studies further demonstrate that localized internal manipulations can have predictable behavioral effects \citep{meng2022rome,minder2024knob}.

Our closest antecedents are intervention-based studies that use internal manipulations to support mechanistic explanations.
The difference here is evidential rather than operational.
Prompt tuning, prefixes, activation steering, editing interventions, and other causal analyses can all reveal useful internal manipulations, but they do not by themselves distinguish between reusing an existing state and rebuilding a new signal elsewhere.
The main novelty of this paper is therefore a stricter evidence standard:
held-out transfer at one fixed interface together with matched necessity, sufficiency, and wrong-interface controls.
The experiments are organized around the settings where that standard can be tested most directly.

\section{What Counts As Evidence}
\label{sec:setup}
We use a schematic write/gate/read vocabulary only to describe the intervention sites studied below.
The important point is simple:
all interface search is done on support data, and the selected intervention is then evaluated once on held-out query data.

\paragraph{Three result types.}
The experiments report three progressively weaker kinds of evidence.
\emph{Single-interface transfer} means that held-out query behavior is preserved by matched replacement at one fixed interface, destroyed by the reverse necessity test, and not reproduced by wrong-interface or random-distribution controls.
\emph{Wider-interface transfer} means that interface interventions remain useful on held-out data but do not satisfy the full single-interface standard, for example because the successful interface must be widened.
\emph{Partial control effects} means that useful internal signal or useful proposals are present, but the decisive controller is distributed or only partly identifiable.

\paragraph{What counts as positive evidence.}
Within a fixed task family, we treat an interface-state variable as stronger evidence than prompt form only when it survives held-out evaluation, wrong-interface negatives, and distribution-matched random controls under support/query lock.
The unit fixed in advance is the interface family and intervention grammar, not one pre-chosen head or neuron.
The pseudo-patch controls are included to test whether success depends on state identity rather than only first- and second-moment matching.
A positive result therefore identifies a strong intervention point within the audited family and intervention grammar.

\paragraph{Locking protocol.}
Support/query lock blocks query-side retuning after support-stage search.
It does \emph{not} remove family choice, budget choice, or intervention-grammar choice.
The appendix records those search axes explicitly.

\paragraph{Recurring checks.}
Across branches we repeatedly ask three simple questions:
\begin{enumerate}
  \item \textbf{Reader-content asymmetry}: \(K\)-only and \(V\)-only interventions should produce different partial drops, while joint \(KV\) replacement should satisfy full necessity/sufficiency at the matched interface.
  \item \textbf{Identity over moments}: distribution-matched random/swap perturbations should fail to close behavior if state identity is broken, even when aggregate energy statistics are matched.
  \item \textbf{Trajectory load growth}: mediator budget and transferable prefix demand should increase with rollout horizon in multi-step generation.
\end{enumerate}
These checks guide interpretation across the tested intervention families.

The empirical program asks four questions, in descending order of strength.
Can a writable interface family yield a localized selected interface?
Can matched state replacement satisfy necessity and sufficiency at that interface?
How much of that single-interface picture survives when multi-step generation requires broader query-side state?
And how does the control picture change once the task becomes reasoning-heavy?
The appendix summarizes the full evidence chain, while the main text follows the shorter routing-first argument.

\section{Experimental Protocol}
\label{sec:protocol}
\begin{table}[tbp]
  \centering
  \scriptsize
  \begin{tabularx}{\textwidth}{@{}L{0.10\textwidth}L{0.16\textwidth}L{0.23\textwidth}L{0.20\textwidth}Y@{}}
    \toprule
    Branch & Support / query unit & Frozen before query & Matched negatives / baselines & Role in the study \\
    \midrule
    Toy & support examples / held-out query examples & write site, inversion key, supervision scope & deeper-layer controls, last-answer-only supervision & inspectable localization and lock sanity \\
    GPT-2 routing & support prompts / query prompts & layer, site, position, head ranking, top-$k$ budget & wrong layer/site, permutation, random-head ablations & main single-interface routing evidence on \routeacc \\
    GPT-2 generation & support rollouts / held-out copy$N$ rollouts & interface scope, \(Q/K/V\) budget & tok1-only transport, swap controls, mean/cov-matched pseudo patches & wider-interface extension on \tokacc\ and \exactacc \\
    Qwen & support split / held-out routing, verify, solve split & interface slot, layer, headset, controller budget & prompt-last, random \(KV\), top1-vs-oracle comparisons & cross-architecture routing consistency check plus reasoning-side slices \\
    \bottomrule
  \end{tabularx}
  \caption{Branch map: what is frozen before query, what the matched comparisons are, and which branches test single-interface transfer, wider-interface transfer, or partial control effects.}
  \label{tab:branch_map}
\end{table}

\paragraph{Support/query lock.}
All interface selection, head ranking, and threshold locking are performed on support subsets; query data are used once for final evaluation.
What is fixed in advance is the interface family and the search procedure, not a single winning slot before support is seen.
Any post-query redesign is treated as a new protocol branch rather than a continuation of the same experiment.

\paragraph{Interventions and controls.}
Interventions follow a nested hierarchy from control-embedding inversion, to compiled residual patch injection, to targeted $Q/K/V$ replacement and attention-edge diagnostics.
Matched controls include permutation, token shift, wrong-layer/site interventions, and layer-matched random head ablations.
For distribution-matched pseudo-patch controls, we use per-token/per-head mean/cov-matched mix controls implemented by non-identity \(K\times K\) orthogonal mixing that preserves the mean axis for each token/head slice.
These controls preserve first/second moments while breaking sample identity, allowing us to test whether transfer depends on state identity rather than coarse activation statistics.

\paragraph{Branches and metrics.}
The toy branch is used for exhaustive localization and protocol validation.
GPT-2 routing is evaluated by \routeacc, while GPT-2 generation is evaluated by both \tokacc\ and \exactacc\ so that token-level and sequence-level transport can be separated.
Qwen is split into routing, verify, and solve settings so that single-interface transfer, state-level reasoning effects, and selector identifiability are not conflated.
A source-locked deployment branch is retained in the appendix because it changes task family, metric, and protocol too substantially to serve as main-text mechanism evidence.

\paragraph{Reporting protocol.}
Each branch includes its main positive result, matched negatives, and unresolved failure mode; when relevant, uncertainty is summarized with confidence intervals and sign-flip tests.
Comparisons are branch-local because residual, $KV$, and $QKV$ interventions expose different internal access, and no single benchmark equalizes access, supervision, and compute across all branches.
The resulting picture is branch-specific: single-interface transfer for GPT-2 routing, the same local routing pattern in Qwen, wider-interface trajectory transport for GPT-2 generation, state-level effects for Qwen verify, and proposal-versus-commit separation for Qwen solve.
Support-stage search burden is reported explicitly because successful selection and unique mechanistic status are different questions.
The appendix records the corresponding search-space and lock decisions, including the appendix deployment branch.

\paragraph{Baselines.}
The baseline suite has two parts.
First, each branch includes direct prompt-form negatives such as wrong-slot or wrong-interface placement; on Qwen routing, for example, the correct operator-token interface is compared directly against the prompt-last control.
Second, the routing branches include matched learned baselines on the same held-out split.
For GPT-2 add/sub and GPT-2 triop, the appendix reports same-budget comparisons among compiled transfer, plain LoRA, LoRA plus trainable control embeddings, and prompt-slot tuning.
In these reruns, the plain LoRA baseline is a deliberately local weight-update baseline applied to the early attention output projection aligned with the retained routing locus; the goal is not to exhaust every PEFT design, but to test whether a small local weight update can replace no-retrain state transfer under the same data and optimization budget.
These baselines are designed to answer two concrete questions: whether a small local weight update in the early attention stack can match no-retrain transfer, and whether trainable prompt slots can rebuild the same behavior elsewhere under the same data and optimization budget.

\paragraph{Reproducibility and lock discipline.}
All model-selection steps are support-locked before query evaluation.
Reported means and standard deviations are across five seeds unless stated otherwise.
The fresh GPT-2 add/sub rerun bundle uses one shared held-out query split across donor inversion, compiled patch transfer, plain LoRA, trainable-control and trainable-slot variants, and the slot-only tuning variants, so those rows are directly comparable rather than stitched from mismatched runs.
Appendix analyses report confidence intervals and sign-flip tests where applicable.
Every positive result is paired with matched negative controls (permutation, token shift, wrong layer/site, and random head baselines).

\section{Routing: State Transfer}
\label{sec:routing_core}
\subsection{Triop at a single early interface}
We start with the strongest positive witness:
GPT-2 triop (add/sub/copy), where the fixed-interface claim can be tested literally at one early site.
At layer-0 block@ctrl, compiled transfer nearly matches donor behavior while staying far above the chance receiver baseline ($0.799\pm0.040$ vs $0.824\pm0.030$ vs $0.333$).
The patch is also intrinsically low-dimensional: at $\delta=0.01$, $r^*=1$ already matches full compiled performance.

\begin{figure}[tbp]
  \centering
  \begin{minipage}{0.49\linewidth}
    \centering
    \includegraphics[width=\linewidth]{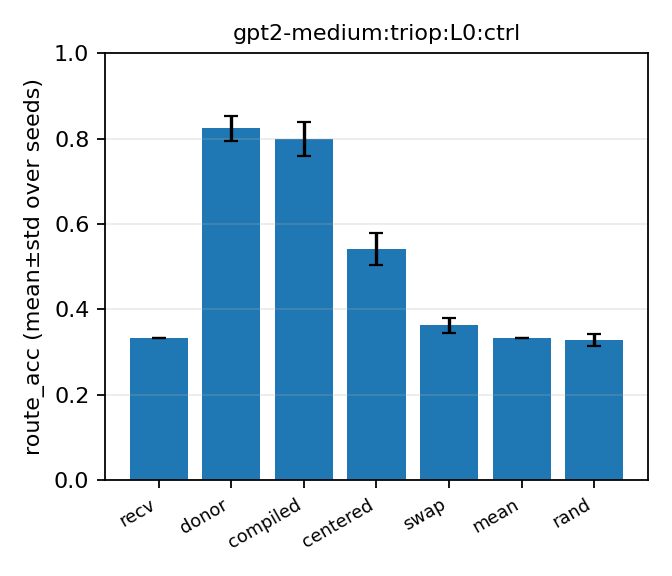}
  \end{minipage}\hfill
  \begin{minipage}{0.49\linewidth}
    \centering
    \includegraphics[width=\linewidth]{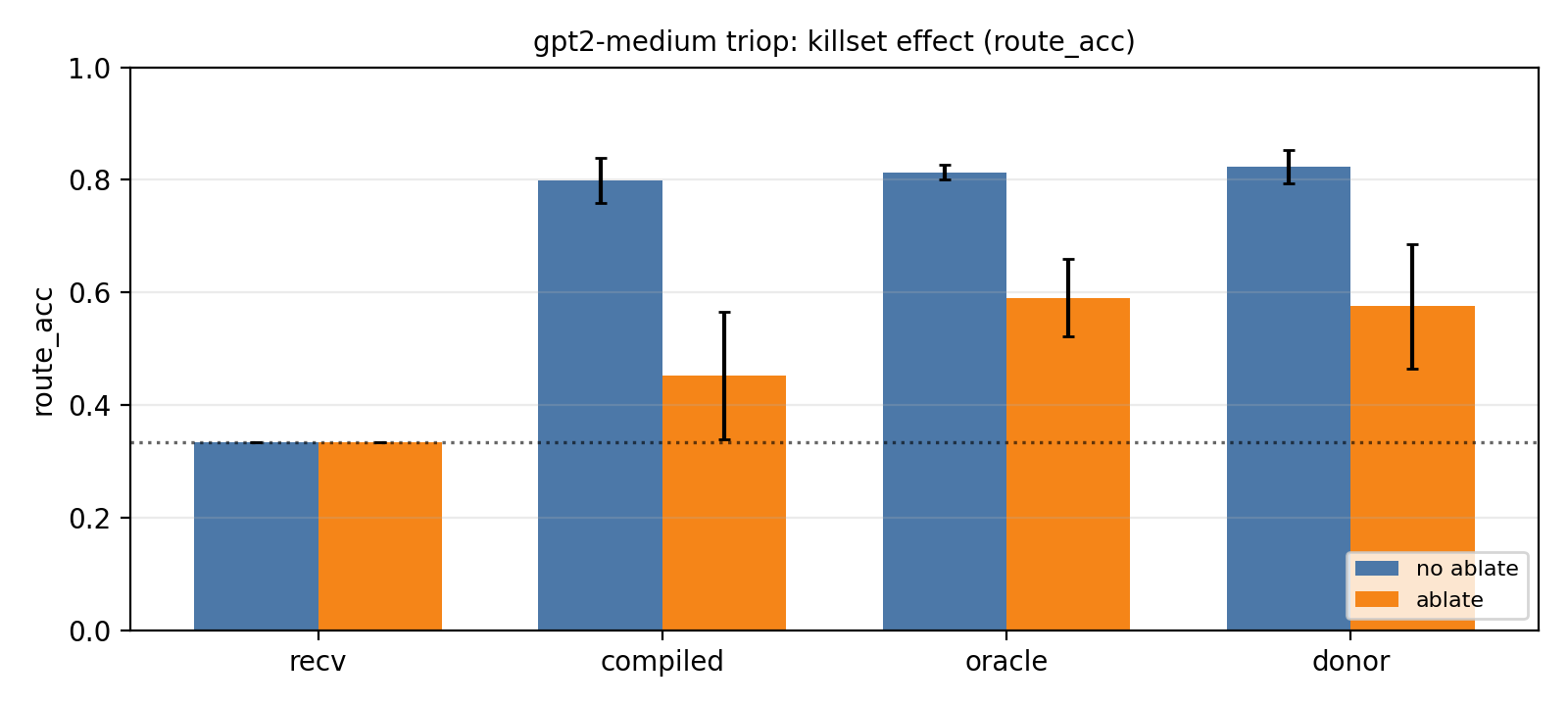}
  \end{minipage}
  \caption{GPT-2 triop: the left panel shows exact transfer at the matched interface; the right panel shows that the strongest structured top-12 gate ablation only partially degrades compiled behavior.}
  \label{fig:triop_core}
\end{figure}

\begin{table}[tbp]
  \centering
  \scriptsize
  \begin{tabular}{@{}L{0.38\linewidth}L{0.23\linewidth}L{0.23\linewidth}@{}}
    \toprule
    Condition (triop, query) & \routeacc\ mean$\pm$std & Interpretation \\
    \midrule
    receiver & $0.333\pm0.000$ & chance baseline \\
    donor & $0.824\pm0.030$ & learned control embedding \\
    compiled & $0.799\pm0.040$ & support-selected patch transfer \\
    centered & $0.541\pm0.038$ & centered-only patch baseline \\
    compiled $K\leftarrow$ centered & $0.723\pm0.046$ & partial necessity (reader targeting) \\
    compiled $V\leftarrow$ centered & $0.611\pm0.027$ & partial necessity (content path) \\
    compiled $KV\leftarrow$ centered & $0.541\pm0.038$ & exact necessity (equals centered) \\
    centered $KV\leftarrow$ compiled & $0.799\pm0.040$ & exact sufficiency (equals compiled) \\
    \bottomrule
  \end{tabular}
  \caption{Triop transfer with split $K$/$V$ necessity diagnostics.}
  \label{tab:triop_closure}
\end{table}

\subsection{Fresh add/sub reruns}
GPT-2 add/sub separates fixed-interface transport from learned relocation.
In a fully local rerun bundle with one shared held-out split summarized in the appendix, donor inversion reaches \(0.9422\pm0.0010\) query \routeacc\ and compiled patch transfer at \(L1@\mathrm{ctrl}\) reaches \(0.9113\pm0.0010\), so direct state transfer recovers most of the donor effect without retraining.
Under the matched 800-step learned baseline budget, plain LoRA on \(L0\) \texttt{attn.c\_proj} remains at chance \(0.5000\pm0.0000\), while LoRA plus trainable control embeddings rises to \(0.9480\pm0.0035\) and control-slot-only tuning reaches \(0.9590\pm0.0028\).
The same pattern holds when the original control rows are frozen random:
LoRA plus trainable \texttt{eq}-slot embeddings reaches \(0.9402\pm0.0109\), and slot-only tuning reaches \(0.9484\pm0.0075\) at \texttt{eq}, \(0.9523\pm0.0035\) at \texttt{a}, and \(0.9547\pm0.0038\) at \texttt{b}.
Once that state is itself trainable, several prompt-slot variants recover the branch at or above donor performance.
The baseline comparison is therefore not limited to one weak alternative.
It shows, on one shared split, that plain local low-rank weight updates fail, while trainable control or prompt slots can eventually match or exceed compiled transfer once nonzero optimization is allowed.

\begin{table}[tbp]
  \centering
  \scriptsize
  \begin{tabular}{@{}lrrrr@{}}
    \toprule
    Slot & support & steps & cost & acc \\
    \midrule
    \texttt{ctrl} & 16 & 200 & 3200 & 0.9128 \\
    \texttt{b} & 16 & 200 & 3200 & 0.9225 \\
    \texttt{a} & 8 & 800 & 6400 & 0.9128 \\
    \texttt{eq} & 32 & 800 & 25600 & 0.9368 \\
    \bottomrule
  \end{tabular}
  \caption{Minimal learned budget required to match or exceed compiled no-retrain transport (\routeacc\ \(=0.9113\)) on the shared GPT-2 add/sub split. Here cost denotes support$\times$steps.}
  \label{tab:addsub_match_compiled}
\end{table}

Appendix Figure~\ref{fig:addsub_compiled_vs_relocation} shows that relocation is regime-dependent rather than uniformly easy.
With only 4 support pairs, wrong-slot tuning remains well below the original control slot even after longer optimization, but by 16--32 support pairs the high-performance basin opens across \texttt{ctrl}/\texttt{eq}/\texttt{a}/\texttt{b}.
Compiled transport is therefore most distinctive in \emph{transport efficiency}, not endpoint accuracy:
on the same held-out split it reaches \(0.9113\) with zero retraining steps, whereas learned relocation needs \(3200\) to \(25600\) support$\times$optimization units to clear the same threshold (Table~\ref{tab:addsub_match_compiled}).
Detailed geometry, attention, and ablation diagnostics move to the appendix; they indicate a shared early access pathway with slot-specific late readout, while compiled patch transfer preserves the sharper fixed-interface causal handle.

\subsection{Localization and specificity}
The same triop interface is sharply localized and mediated by a small but non-singleton head set.
Performance is best at layer 0 and degrades monotonically to chance by layer 23, while only ctrl-site writes remain strong at the preferred depth (block@ctrl $0.799$, mlp@ctrl $0.789$, attn@ctrl $0.344$).
Ranked top-$k$ ablations are far more destructive than layer-matched random baselines, but collapse only at large $k$, which points to concentrated mediation with distributed redundancy rather than a single decisive head.

\begin{figure}[tbp]
  \centering
  \begin{minipage}{0.49\linewidth}
    \centering
    \includegraphics[width=\linewidth]{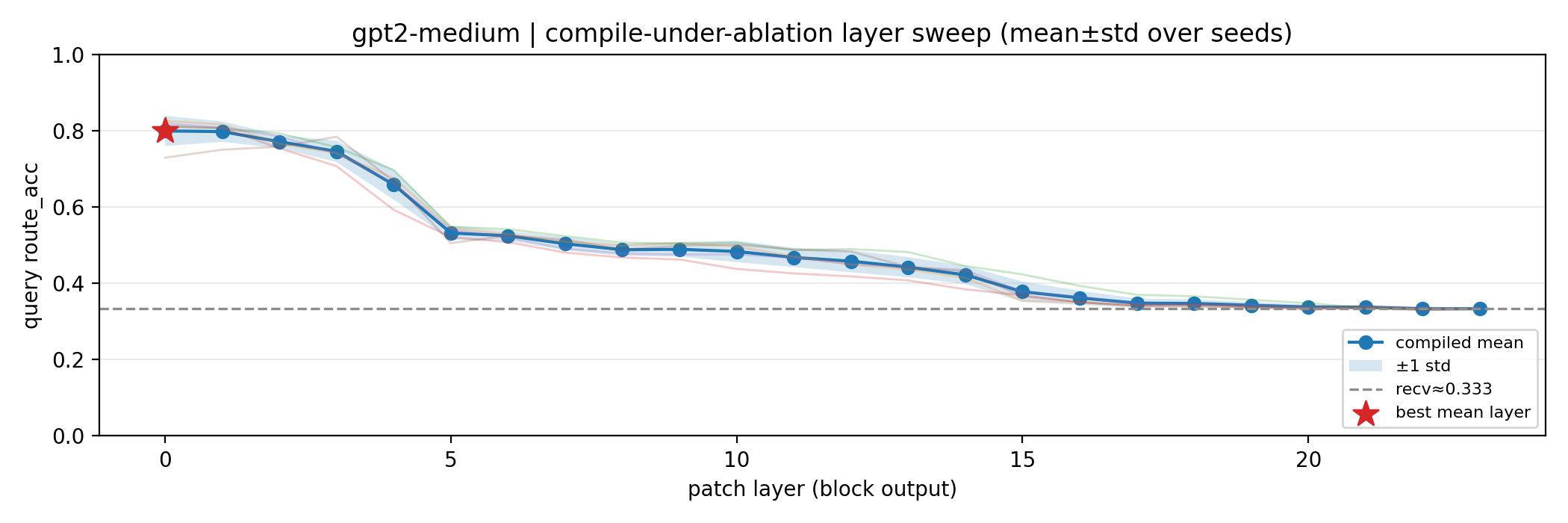}
  \end{minipage}\hfill
  \begin{minipage}{0.49\linewidth}
    \centering
    \includegraphics[width=\linewidth]{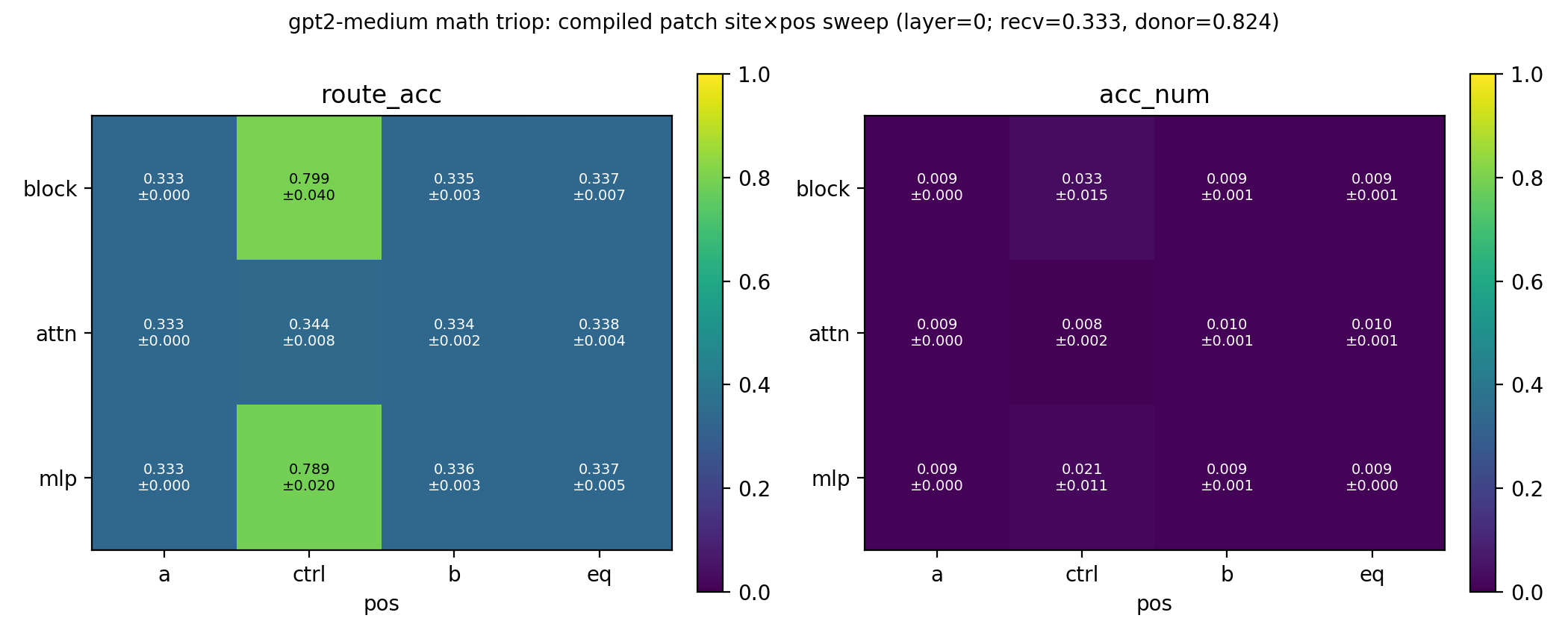}
  \end{minipage}
  \caption{Triop localization: early layers dominate (left), and control position is privileged (right).}
  \label{fig:triop_localization}
\end{figure}

Matched-random controls with the same layer profile and head budget do not reproduce the selected mediator effect, so routing specificity depends on head identity rather than ablation energy.
In the locked randomization test, the selected 12-head set is more destructive than every matched random set (\(\Delta\routeacc=-0.0402\) vs random mean \(-0.0042\pm0.0060\), range \([-0.0148,+0.0072]\); \(n=60\)).
Support-ranked top-$40$ heads already recover most compiled behavior (\(0.757\pm0.033\)), and compiled route accuracy remains high across prompt styles (\texttt{math} to \texttt{calc}: \(0.799\pm0.040 \rightarrow 0.783\pm0.041\)).
The appendix summarizes the routing-side evidence chain, while the remaining mediator-budget and specificity diagnostics are omitted from the main text for brevity.

\subsection{Toy support for localization and lock discipline}
The toy branch provides a compact sanity check on the same early-interface story.
Patch-position sweeps localize the writable site sharply to an early residual location centered on \texttt{b\_last}: at threshold $0.9$, layer-0 yields $45$ minimal successful mediator sets while layer-1 yields none, and the token group containing \texttt{b\_last} appears in $97.8\%$ of minimal layer-0 solutions.

Mechanism-aligned residual compilation then identifies a minimal early interface sufficient for execution: at layer-0 \texttt{ctrl+b\_last}, compiled interventions reach \(0.349\pm0.011\), while \texttt{ctrl}-only and \texttt{b\_last}-only variants are much weaker and matched layer-1 interventions collapse to the discrete baseline.
Support-only key selection also reproduces oracle-style query selection up to tiny residual gaps (\(\Delta \le 0.007\)), so the toy branch mainly validates localization and lock discipline rather than adding a separate main-text claim.

\section{Generation Requires Broader Interfaces}
\label{sec:generation}
\subsection{Interface composition in copy2 generation}
Copy2 generation is the clearest boundary case outside routing.
It supports trajectory transport under a broader prompt-conditioned interface than the one-site routing results.
As the writable/readable interface expands from a narrow ctrl-only site to broader prompt-wide state, exact-match transport rises sharply; matched attention-only and MLP-only alternatives remain weak or fail outright.
Multi-step control is therefore distributed across a broader prompt-conditioned interface rather than closing at one small routing-style site.

\begin{figure}[tbp]
  \centering
  \includegraphics[width=0.72\linewidth]{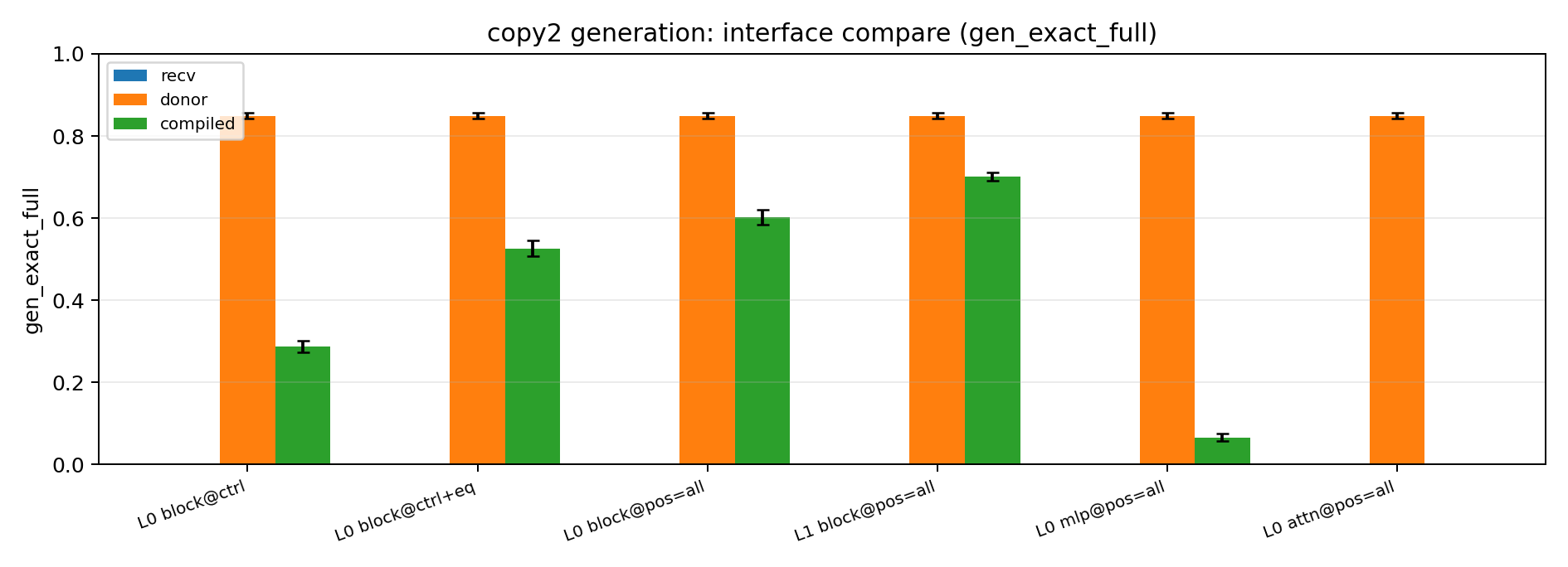}
  \caption{Copy2 generation interface comparison. Transport improves as the write/read scope becomes broader.}
  \label{fig:copy2_interface}
\end{figure}

\subsection[From local KV transport to broader QKV transport]{From local \(KV\) transport to broader \(Q\times KV\) transport}
In a stricter gate-composition protocol, copy2 generation shows a clear assembly pattern.
Necessity is strong: replacing compiled \(KV(\texttt{ctrl})\) by centered values collapses \exactacc\ close to centered baseline.
Sufficiency is initially partial, then improves monotonically as the interface is expanded from local \(KV\) transport to prompt-wide \(KV\) transport and finally to \(Q\times KV\) transport that includes query-side alignment.
This sharpens the interpretation: multi-token generation control is not ``\(KV\)-only''; it is a compositional \(Q\times KV\) interface where query-side alignment becomes increasingly important.
The branch therefore supports partial trajectory transport with a wider interface rather than the clean single-interface result seen in routing.

\subsection{QKV versus KV across copyN}
Across $N=\{2,3,5,8,10\}$, centered$\leftarrow QKV$ outperforms centered$\leftarrow KV$ in \tokacc.
The gap is persistent rather than isolated to one horizon; for example, on copy10, adding query-side state closes much of the distance between \(KV\)-only transport and compiled behavior.
Thus query-side state is part of the transportable interface in generation rather than a secondary detail.

\begin{table}[tbp]
  \centering
  \small
  \begin{tabular}{@{}c c c c@{}}
    \toprule
    $N$ & compiled & $c\leftarrow KV$ & $c\leftarrow QKV$ \\
    \midrule
    2  & 0.855 & 0.727 & 0.817 \\
    3  & 0.772 & 0.586 & 0.720 \\
    5  & 0.661 & 0.447 & 0.598 \\
    8  & 0.671 & 0.441 & 0.581 \\
    10 & 0.700 & 0.504 & 0.650 \\
    \bottomrule
  \end{tabular}
  \caption{Copy$N$ generation scaling in \tokacc\ (here $c$ denotes centered baseline state).}
  \label{tab:copyN_qkv}
\end{table}

\begin{figure}[tbp]
  \centering
  \includegraphics[width=\linewidth]{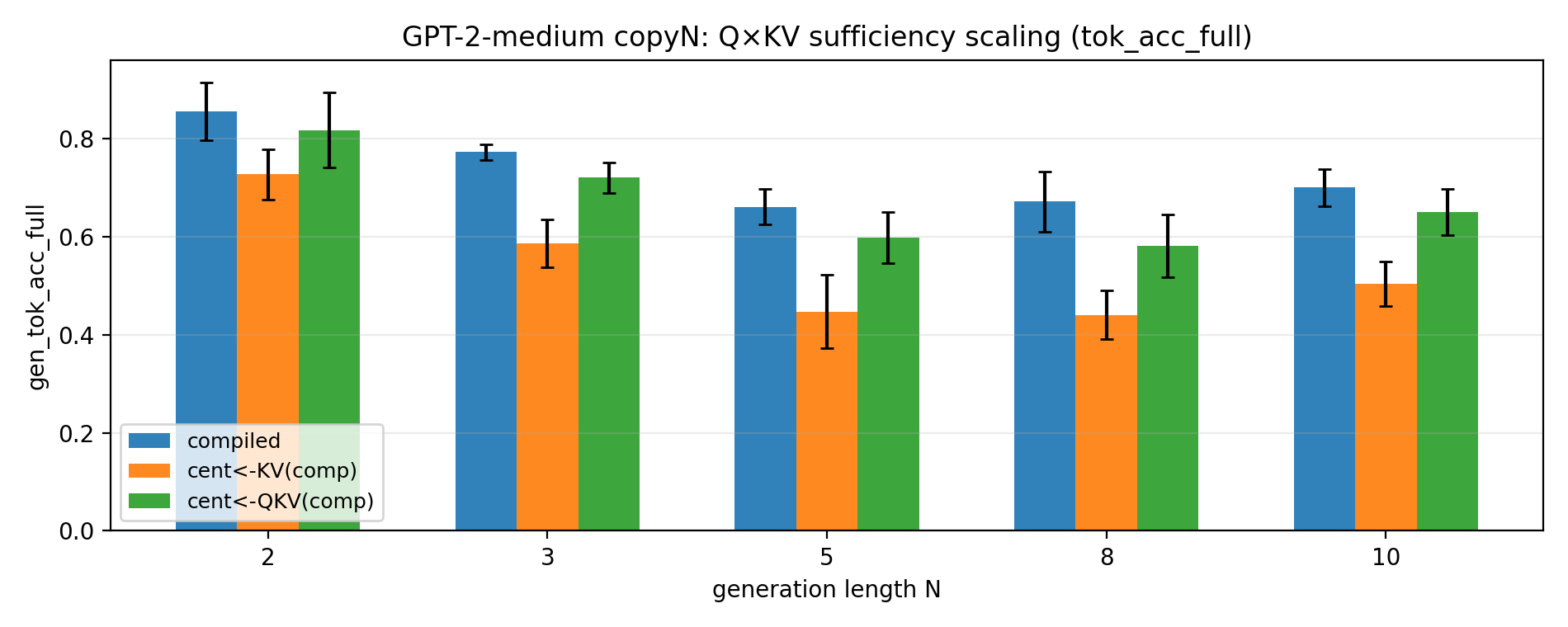}
  \caption{Copy$N$ generation scaling for interface interventions.}
  \label{fig:copyN_qkv}
\end{figure}

\subsection{Fixed-corrupt tests: tokpos vs tok1}
We corrupt visible prompt content and evaluate state transport directly.
On corrupt-only scoring, tokpos transport remains strong while tok1-only transport collapses across horizons.
This favors trajectory-state transport over tok1-only transport.

Under an additional copy10 swap-adjacent stress test, replacement from compiled states remains positive while distribution-preserving cross-example swaps collapse, and the replace-minus-swap gap stays positive in all seeds.
Thus second-order structure alone does not close behavior: transport depends on task-aligned state identity, not only broad activation statistics.
Together with the remaining compiled--transport gap in \Cref{tab:copyN_qkv}, these checks support trajectory-state transport under a broader interface.

\begin{figure}[tbp]
  \centering
  \includegraphics[width=\linewidth]{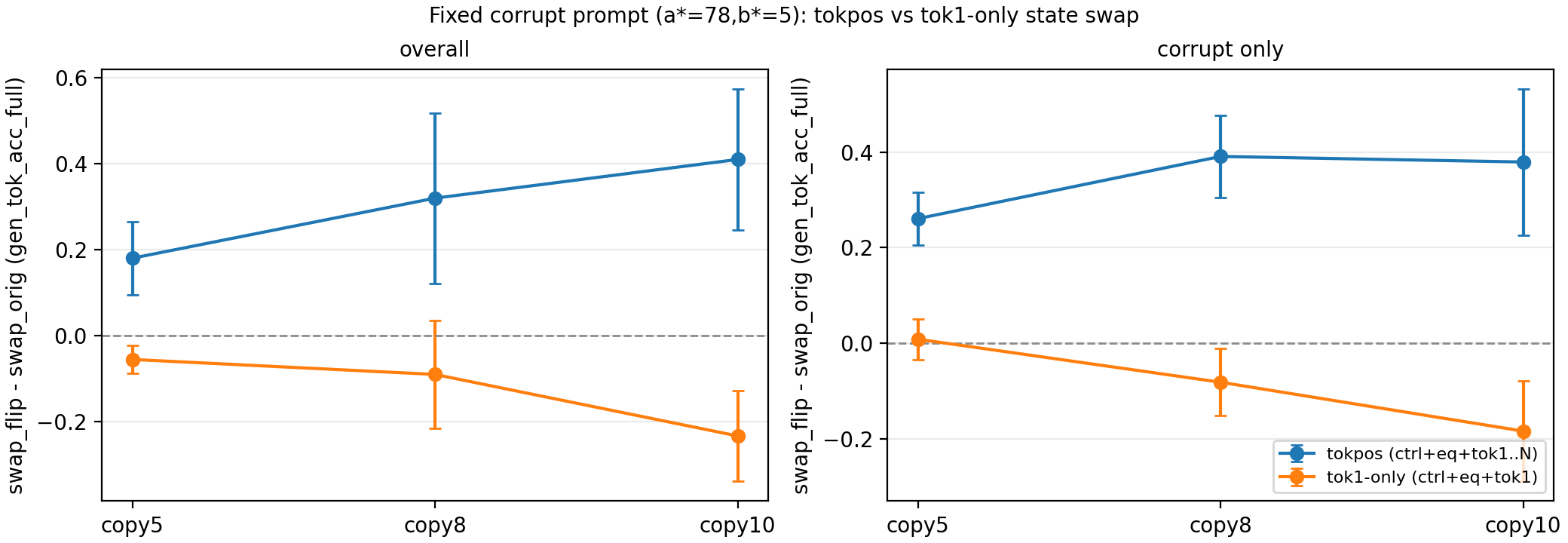}
  \caption{Fixed-corrupt protocol: trajectory-level state transport remains effective; tok1-only does not.}
  \label{fig:fixed_corrupt}
\end{figure}

Additional generation diagnostics, including rollout forecastability and prompt$\rightarrow$state inversion, support the same broader-interface reading but are secondary to the main routing result.

\section{Qwen Routing and Reasoning: Scope and Limits}
\label{sec:qwen}
We next test whether the routing result survives in a different architecture and task family.
This section contributes one cross-architecture routing consistency check together with two reasoning-side extensions.
The routing branch tests whether the same matched-interface pattern reappears in a second architecture, while the verify and solve branches show what remains once control depends on a broader or harder selection problem.

\subsection{Routing at the operator token}
Qwen recovers the matched-interface routing pattern at the operator token.
Compiled state injection reaches perfect routing, while the matched prompt-last intervention fails badly; receiver$\leftarrow$donor and donor$\leftarrow$receiver swaps invert behavior in opposite directions.
The content path is sufficient and specific:
\(V\)-side replacement closes routing, \(K\)-only replacement does not, and matched random \(KV\) replacement stays near control levels.
The same split survives stricter controls.
With the prompt shell shared, \(V\)-only replacement still closes routing at \(1.0\), while \(K\)-only and same-prompt \(KV\) negatives remain at \(0.1875\).
In a local Qwen2.5-3B rerun with a fixed receiver shell, \(V\)-only transfer still closes routing and \(K\)-only still fails.
Lexical remaps to \texttt{add/sub}, \texttt{plus/minus}, and \texttt{Task: ... plus/minus ...} also preserve compiled and \(V\)-only transfer at \(1.0\).

\begin{figure}[tbp]
  \centering
  \includegraphics[width=0.90\linewidth]{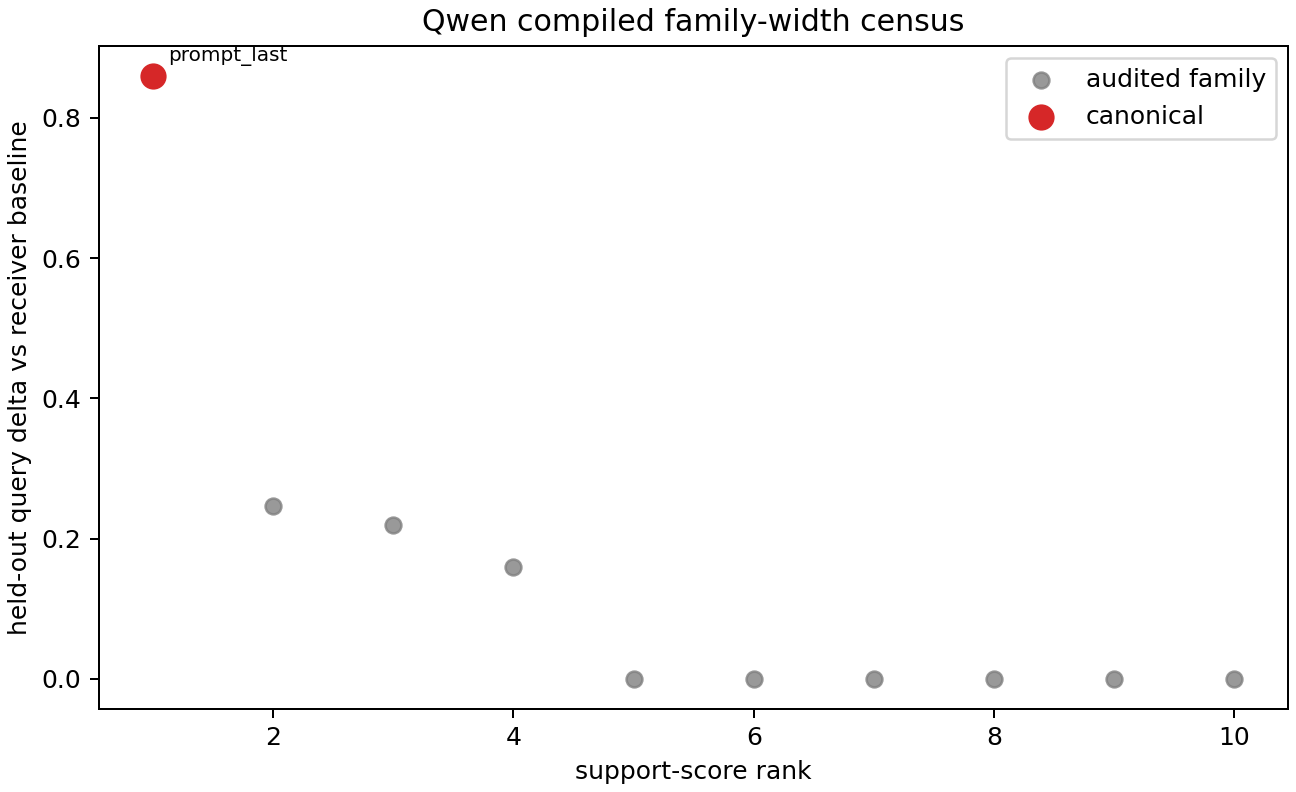}
  \caption{Qwen local compiled family-width census under matched support-selected search. Canonical \texttt{prompt\_last} stays rank-1 on both support and held-out query, and no alternative local mode lands within \(0.05\) of its held-out rescue score.}
  \label{fig:qwen_compiled_family_census}
\end{figure}

The remaining uncertainty is central, not peripheral: donor-specific identity on the local \(V\)-path is unresolved in the tested family.
In the receiver-patched Qwen2.5 rerun, matched random \(KV\) rises to \(0.863\), and the harder remap series lowers random \(KV\) from \(0.852\) to \(0.664\) to \(0.520\) and then \(0.488\) without changing the aligned \(V\)-only result.
Altbank donor-answer checks and the five-seed instance-binding benchmark point the same way:
compiled routing remains perfect, but aligned and permuted \(V\)-only transfer do not separate cleanly, and donor-answer preference stays near chance.
Within the local family tested here, the effect is nevertheless sharply site-selective.
Figure~\ref{fig:qwen_compiled_family_census} shows a ten-mode local census selected on support data in which canonical \texttt{prompt\_last} stays rank-1 on both support and held-out query (\(0.8752\) and \(0.8603\)), the strongest alternative \texttt{prompt\_last\_minus\_4} reaches only \(0.2615\) support and \(0.2468\) query, and no alternative lands within \(0.05\) of the canonical held-out score.
The Qwen result is therefore best read as cross-architecture evidence compatible with the routing account, not as a clean identity-level replication of the GPT-2 result.

\subsection{Verify: state-level gain under strict holdout}
Qwen verify shows a robust state-level effect under strict holdout.
On strict holdout, state-compiled no-CoT\(+\delta\) consistently beats both no-CoT and CoT prompts, with positive deltas in all \(82/82\) evaluation runs.
Matched-random ablation and remove\(+\)restore tests again point to a localized mediator, but the branch does not exhibit the routing-style swap logic or one-interface necessity/sufficiency pattern.
This branch therefore isolates a useful mediator without closing the full routing-style transfer loop.

\subsection{Solve: proposal-rich but commit-limited control}
Solve separates proposal quality from committed selection.
Across two cohorts, oracle shortlists recover most of the available gain with small branching budgets, while the best learned controller remains near the one-proposal regime.
In the retained seed25--29 analysis, full oracle gain is \(+0.422\), oracle@\(B=4\) already recovers \(+0.372\), oracle@\(B=8\) reaches \(+0.418\), and the best learned controller reaches only \(+0.068\).
Matched \(k\)-sweeps tell the same story:
substantial proposal signal survives, but robust committed selection and reranking do not.

\begin{figure}[tbp]
  \centering
  \includegraphics[width=0.94\textwidth]{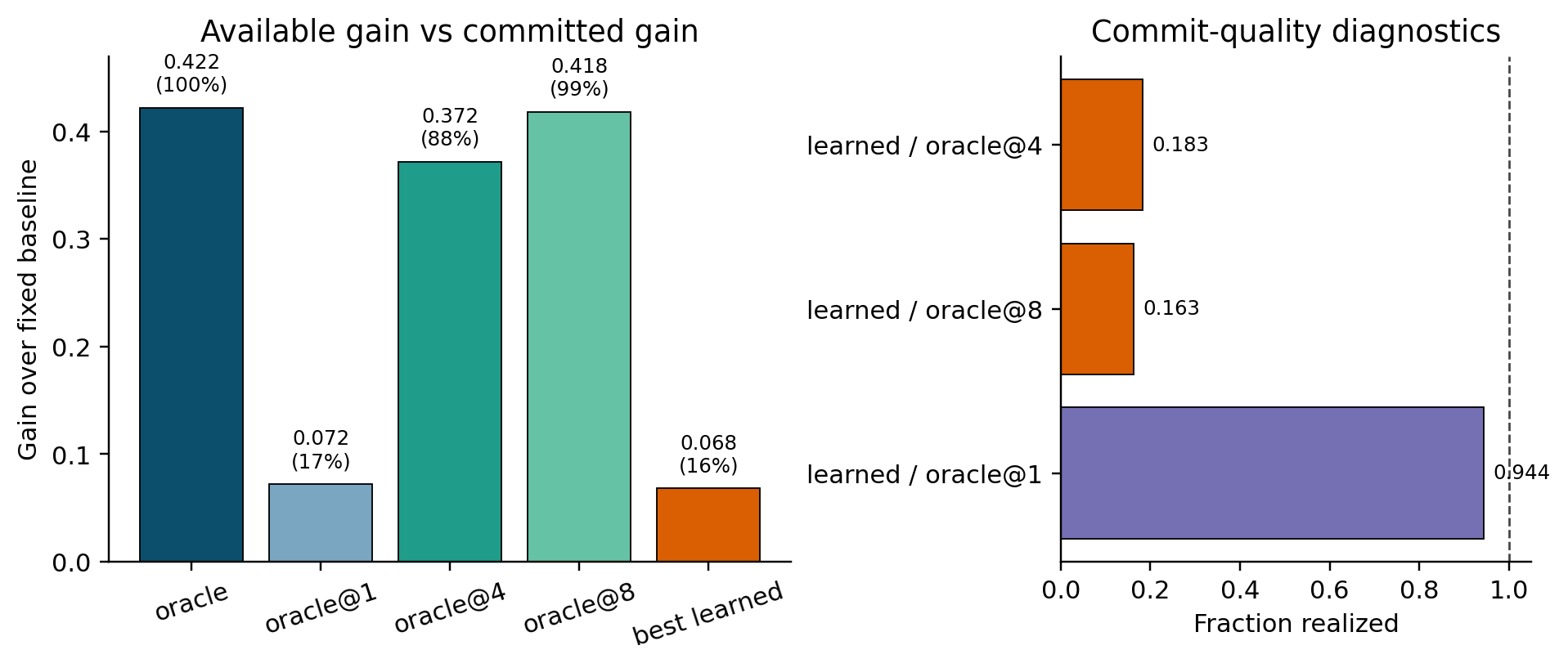}
  \caption{Within the retained controller comparison, Qwen solve is proposal-rich but commit-poor. Small oracle shortlists already recover most of the available gain, while the best retained learned controller remains near the oracle@1 regime instead of behaving like an effective reranker over a high-quality shortlist.}
  \label{fig:qwen_solve_proposal_commit_gap}
\end{figure}

Solve therefore exposes a separation between proposal quality and committed selection rather than a routing-style controller.

A compact numerical summary of the Qwen routing, verify, and solve branches is given in the appendix.
That appendix also includes one frozen-grammar cross-task audit: with the same fixed router grammar on \texttt{ifeval} and \texttt{beavertails}, compiled transfer remains positive but recovers only a small fraction of donor routing (\(0.0327\) and \(0.1037\)).

\section{Discussion}
\label{sec:discussion}
The main result is that controlled routing can be tracked by transferable internal state more sharply than by prompt form alone.
Toy localization identifies writable early interfaces, GPT-2 triop gives exact single-interface transfer, and GPT-2 add/sub separates no-retrain transfer from learned relocation.
These routing results establish the paper's strongest claim: within an audited interface family, fixed-interface transfer gives stronger evidence for reuse than trained prompt success alone.

The add/sub branch makes the central comparison concrete.
On the same held-out split, donor control inversion reaches \(0.9422\pm0.0010\), compiled patch transfer reaches \(0.9113\pm0.0010\), and plain LoRA stays at chance.
Once prompt slots are trainable, the same endpoint can be relearned broadly: control-slot tuning, non-control-slot tuning, and slot-only tuning at \texttt{eq}/\texttt{a}/\texttt{b} all recover high routing accuracy.
Compiled transport is distinctive because it reaches the target behavior without retraining and with much lower adaptation cost: learned baselines that match the compiled score band require \(3200\) to \(25600\) support\(\times\)steps on add/sub, \(25600\) on triop, and \(6400\) on Qwen local routing.
This is the empirical separation between reuse and relocation that motivates the paper.

The same comparison now appears, in weaker but still useful form, on triop.
Under an equalized 800-step budget, compiled transfer reaches \(0.7979\), plain LoRA remains at chance \(0.3336\), prompt-slot tuning reaches only \(0.6818\) to \(0.7089\), and LoRA plus trainable control embeddings reaches \(0.8036\).
So the routing story is not add/sub-only: in both retained GPT-2 routing branches, plain local weight updates fail, while learned control states can catch up only with optimization.

Qwen routing is best interpreted as a consistency check on that picture rather than a second strongest-form closure result.
It recovers the same matched-interface pattern and strong site selectivity in a second architecture, but donor-specific identity on the local \(V\)-path remains unresolved.
That makes the cross-architecture evidence supportive but incomplete.

The later branches mainly delineate scope rather than strengthen the main closure claim.
Generation still supports transfer, but the useful interface widens from a localized routing site to broader \(Q\times KV\) trajectory state.
Qwen verify retains a localized state-level gain, while Qwen solve shows that strong proposals do not by themselves yield strong committed control.
An intermediate cross-task Qwen audit sits between these cases: under one frozen grammar, compiled gains remain positive across held-out benchmark families, but donor recovery remains low.
These later branches therefore map how the routing account degrades as task family, trajectory width, and controller difficulty grow.

\section{Limitations and Scope}
Four limitations are most important.
First, support-stage selection still ranges over predefined interface families and search procedures rather than a single globally fixed site.
The resulting claim is about a support-selected reusable interface within the audited family, not about a unique canonical site across the whole model.

Second, the search burden is substantial.
The appendix shows that one target11 slice contains \(8896\) candidates at the locked source target, that \(252\) perfect-source rules already tie at a \(29/32\) external ceiling in a simpler family, and that \(2880/5760\) source-only selector orders are zero-regret on the scanned core pools.
A larger boundaryaware scan with \(30140\) rule evaluations still leaves a one-hit residual gap to the slice ceiling.
This multiplicity leaves a real gap between finding an effective interface and establishing a uniquely privileged mechanism.

Third, the strongest evidence comes from controlled and partly algorithmic tasks, with most exhaustive causal grids on GPT-2-medium rather than frontier-scale models.
That choice improves identifiability but narrows ecological breadth.
The paper therefore makes its strongest contribution as a rigorous mechanistic study of routing-style control, rather than as a general account of prompt control in open-ended language tasks.

Fourth, several branches remain only partially characterized.
Some long-horizon generation transfers are incomplete, the pseudo-patch suite does not yet fully separate identity on \(V\)-path content channels, and Qwen solve shows that a large oracle ceiling does not by itself yield an identifiable controller.
On Qwen routing, same-prompt, receiver-patched, alias-transfer, altbank, and instance-binding analyses all support matched-interface transfer, but donor-specific identity on the local \(V\)-path remains unresolved.
When the local family widens to dynamic options with per-example answer grammar, no compiled local mode recovers routing, and the frozen-grammar cross-task results show protocol reuse without routing-level recovery.

Baseline coverage also remains targeted rather than exhaustive.
The current set of baselines includes prompt-form negatives, wrong-slot controls, fixed-direction steering, distribution-matched pseudo-patch controls, matched same-budget reruns for GPT-2 add/sub and triop, and fresh Qwen local-routing reruns.
That is enough to ground the routing-side transport-versus-relocation contrast directly in the retained routing branches, but it does not cover every possible PEFT insertion point, adapter design, or Qwen reasoning baseline.

\section{Conclusion}
This study shows that prompt-induced routing behavior can be identified by fixed-interface state transfer before any retraining is allowed to relocate the control signal.
Early-interface localization, exact single-interface transfer on GPT-2 triop, and the add/sub contrast between no-retrain transport and learned relocation provide the strongest evidence.
Qwen routing adds a useful cross-architecture consistency check on the same pattern.
Taken together, these results support one main conclusion:
for routing-style control, fixed-interface transfer is stronger evidence of reuse than trained prompt success alone.
Generation and reasoning extensions then mark where the same methodology continues to reveal transferable state and where control becomes broader or harder to identify.
The methodological lesson is that localized interface tests, matched negatives, held-out evaluation, and zero-retrain transfer can turn prompt success into a more specific mechanistic claim within a clearly delimited regime.

\bibliographystyle{plainnat}
\bibliography{refs}

\clearpage
\appendix
\section{Appendix}
The appendix keeps only the material needed to audit the main routing claims.
This appendix gives the retained summary of the evidence, the support/query lock, and the source-lock search burden without relying on dense audit tables.
The remaining material extends the routing baseline audit, collects the compact Qwen summaries, and records one frozen-grammar cross-task readout beyond the core routing family.

\FloatBarrier
\subsection{Evidence Summary}
This section restates the retained evidence hierarchy in the same routing-first order as the main text, but with the main numerical anchors collected in one place.

\subsubsection{Localization and Single-Interface Transfer}
Toy early-layer minimal sets and the triop block@ctrl peak provide the retained localization witness.
Deeper-layer and wrong-site controls collapse, so writable interfaces exist and can be support-selected sharply.
Numerically, the main triop anchor remains the layer-0 block@ctrl result: donor reaches \(0.824\pm0.030\), compiled reaches \(0.799\pm0.040\), and the chance receiver baseline is \(0.333\).
This is the cleanest retained example of support-selected transfer at one early interface.

The strongest retained routing result is the triop closure pattern:
\begin{itemize}
  \item centered \(KV\leftarrow\) compiled fully matches the compiled interface;
  \item compiled \(KV\leftarrow\) centered drops back down;
  \item split \(K\)-only and \(V\)-only replacements remain partial.
\end{itemize}
More concretely, compiled \(K\leftarrow\) centered yields \(0.723\pm0.046\), compiled \(V\leftarrow\) centered yields \(0.611\pm0.027\), compiled \(KV\leftarrow\) centered returns to centered at \(0.541\pm0.038\), and centered \(KV\leftarrow\) compiled returns to compiled at \(0.799\pm0.040\).
The retained inference is therefore that gate and content roles are separable but jointly required at one matched interface.

\subsubsection{Cross-Architecture Routing}
Qwen routing retains perfect matched-interface transfer with swappability and wrong-interface failure at the tested site.
The cleanest retained numbers are the operator-token rescue at \(1.0\), the wrong-interface prompt-last control at \(0.383\), and the split result in which \(V\)-only remains at \(1.0\), \(K\)-only stays at \(0.0\), and matched random remains low in the main local split.
The key caveat is unchanged: stronger \(V\)-path checks still leave donor-specific content identity unresolved, and in receiver-patched reruns random \(KV\) can rise well above the routing null.
The Qwen branch therefore functions as a cross-architecture consistency check rather than a strongest-form replication.

\subsubsection{Generation and Reasoning Boundaries}
In generation, the retained point is not one extra closure result but a change in interface width.
Fixed-corrupt tokpos transport remains positive across longer horizons, tok1-only transport fails, and compiled behavior stays above centered\(\leftarrow QKV\).
Across copy\(N\), centered\(\leftarrow QKV\) consistently outperforms centered\(\leftarrow KV\), for example on copy10 where compiled is \(0.700\), centered\(\leftarrow KV\) is \(0.504\), and centered\(\leftarrow QKV\) rises to \(0.650\).
The retained reading is that generation supports partial trajectory transport under a broader \(Q\times KV\) interface.
Within this broader-interface regime, a fixed-direction steering baseline remains weak.
On copy12, the retained layer-0 eq-axis steering baseline reaches \(\tokacc=0.6899\) and \(\exactacc=0.5273\), while the best of five matched random axes averages \(\tokacc=0.6935\pm0.0035\).
So wider-interface transport is not reproduced by an arbitrary fixed steering direction.

For reasoning-side settings, the retained picture is:
\begin{itemize}
  \item Qwen verify shows a robust holdout effect across all evaluation runs;
  \item Qwen solve remains top1-fragile despite a large oracle ceiling.
\end{itemize}
On verify, the main retained gains are \(\Delta_{\texttt{patch-no}}=+0.1530\) and \(\Delta_{\texttt{patch-cot}}=+0.1847\), with positive deltas in all \(82/82\) runs.
On solve, full oracle gain is \(+0.422\), oracle@4 already recovers \(+0.372\), and the best learned controller reaches only \(+0.068\).
These branches show that state-level reasoning effects exist, but identifiability and single-interface transfer are weaker than in routing.

\FloatBarrier
\subsection{Support-Query Lock and Search Burden}
This section spells out what the lock protocol does and does not fix, and why the multiplicity issue remains scientifically relevant even under support/query discipline.

\subsubsection{What Is Locked Before Query}
Interface localization, mediator budget, copy\(N\) forecast rule, onset thresholds, deployment risk weight, source-size mixture, and rule-class selector are all fixed at support stage or source-only support stage.
Query data are reserved for final evaluation plus uncertainty reporting.
The important retained point is procedural: query examples are not used to retune the selected interface, threshold, or controller after support-stage search has completed.

The locked search ranges include:
\begin{itemize}
  \item layers \(\times\) sites \(\times\) positions for interface localization;
  \item top-\(k\) and headset sweeps for mediator budget;
  \item model-family and threshold choices for copy\(N\) forecast rules;
  \item \(\tau\)-grids for onset thresholds;
  \item a \(1001\)-point \(\lambda\) scan for deployment risk;
  \item a \(5151\)-point simplex scan for source-size mixtures;
  \item up to \(13230\) constraint-selector evaluations for rule-class selection.
\end{itemize}

\subsubsection{What Multiplicity Remains}
Source-only locking still leaves a broad admissible pool.
The main retained multiplicity facts are:
\begin{itemize}
  \item For one locked source target, target11 leaves \(8896\) candidates and blind external score spans \(23\) to \(31\) out of \(32\), with median \(27.0\).
  \item In a simpler rule family, exact source fit still tops out at \(29/32\), with \(252\) perfect-source rules tied at that ceiling.
  \item Selector-family tie-breaks remain highly non-unique: \(2880/5760\) source-only selector orders achieve zero regret on the two core pools, and the best scanned order still selects a \(31/32\) rule on both.
  \item Broader search reduces but does not remove the residual gap: one target11 branch scans \(30140\) rule evaluations across \(12294\) candidates, \(1507\) constraints, and \(20\) selectors, yet the best locked rule class still stops at \(30/32\) against a slice ceiling of \(31/32\).
  \item Family choice matters more than raw pool size, since \texttt{chatgptjb1} anchor bundles reach ceiling with \(325\)-candidate pools while matched \texttt{sashas1} bundles remain \(2\) hits short despite \(1488\)-candidate pools.
\end{itemize}
Within the retained deployment slice, one calibrated transfer cube still closes: locked \(\lambda_t=0.322\) is oracle-equivalent on \(9/9\) audited cells.
Taken together, these numbers make the intended appendix point precise: support/query lock rules out post-query retuning, but it does not collapse the scanned family to a unique privileged mechanism.

\FloatBarrier
\subsection{Matched Learned Baselines in Routing}
This section collects the learned baselines that are directly comparable to the retained routing witnesses.
The main pattern is shared across the routing branches: compiled transfer is the only zero-retraining row, plain local weight updates do not recover the branch, and learned relocation succeeds only once trainable control state and optimization are allowed.

\subsubsection{GPT-2 Add/Sub}
This block keeps only the add/sub objects needed for the transport-versus-relocation contrast.
On the shared held-out split, support-locked control inversion remains the clearest learned donor witness at \routeacc\ \(0.9422\pm0.0010\), while compiled patch transfer at \(L1@\mathrm{ctrl}\) reaches \(0.9113\pm0.0010\) and stays far above the receiver baseline.
Plain LoRA on \(L0\) \texttt{attn.c\_proj} remains at chance, \routeacc\ \(0.5000\pm0.0000\), so weight-only low-rank updates on the retained early attention projection do not explain the routing effect in this branch.
Once the original control state is trainable, the learned baselines fully close the branch:
\begin{itemize}
  \item LoRA plus trainable control embeddings reaches \(0.9480\pm0.0035\);
  \item \texttt{ctrl}-slot-only tuning reaches \(0.9590\pm0.0028\);
  \item non-control prompt-slot relearning at \texttt{eq}/\texttt{a}/\texttt{b} ranges from \(0.9402\) to \(0.9547\).
\end{itemize}
The cost comparison is the retained point.
To match or exceed the compiled no-retrain target of \(0.9113\), learned relocation requires \(3200\) support\(\times\)steps at \texttt{ctrl} or \texttt{b}, \(6400\) at \texttt{a}, and \(25600\) at \texttt{eq}.
The retained contrast is therefore not endpoint impossibility but no-retrain transport versus learned relocation cost.

\begin{figure}[tbp]
  \centering
  \includegraphics[width=\linewidth]{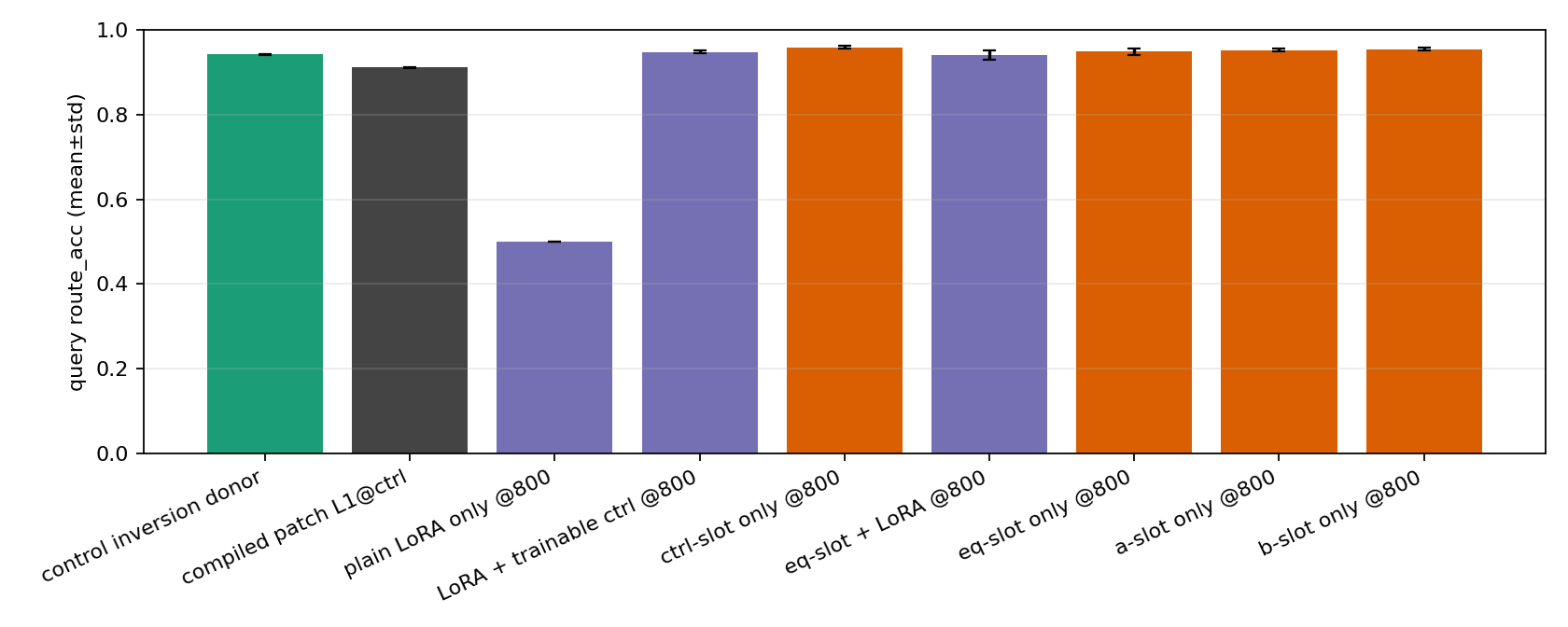}
  \caption{GPT-2 add/sub same-budget learned baseline suite. All learned rows use the same held-out split and an equalized 800-step budget; compiled remains the only zero-retraining row.}
  \label{fig:addsub_samebudget_baselines}
\end{figure}

\begin{figure}[tbp]
  \centering
  \includegraphics[width=\linewidth]{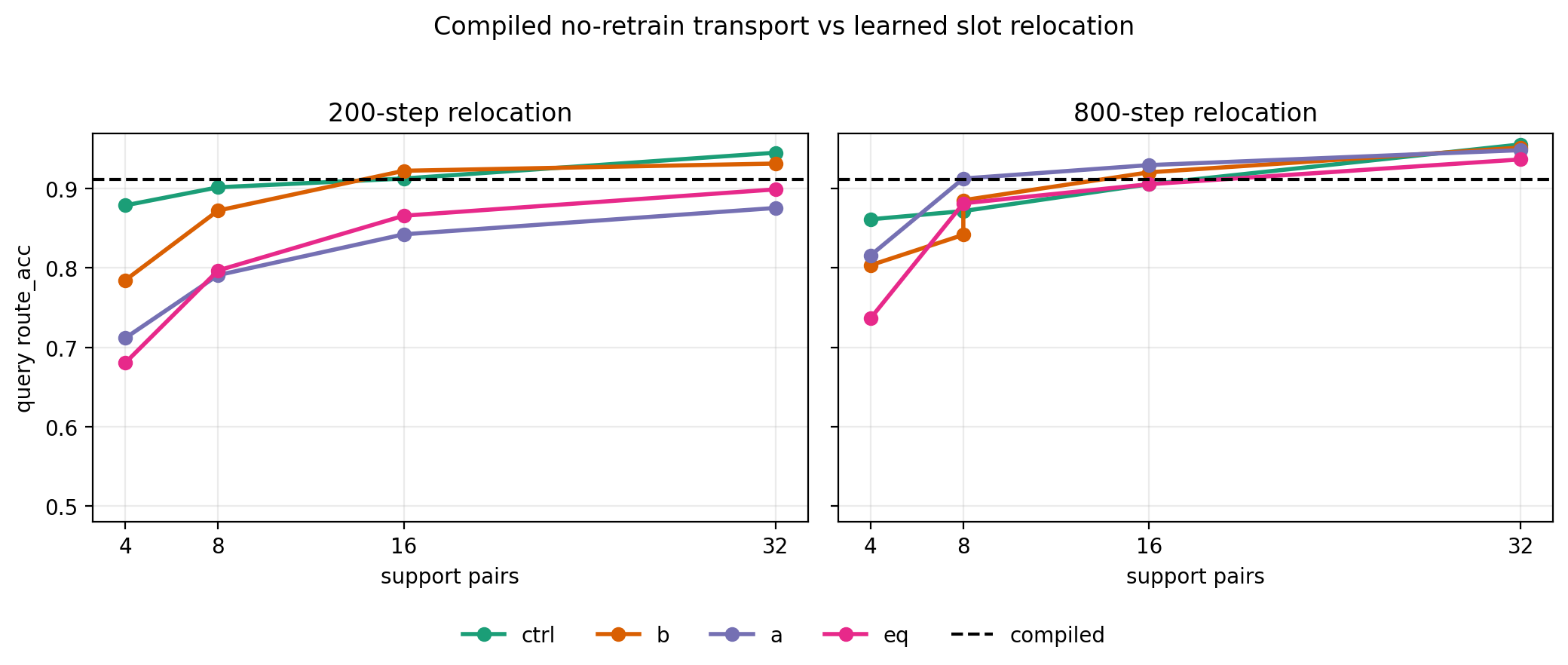}
  \caption{Compiled transport versus learned relocation on GPT-2 add/sub. The dashed line marks the compiled no-retrain score.}
  \label{fig:addsub_compiled_vs_relocation}
\end{figure}

\FloatBarrier
\subsubsection{GPT-2 Triop}
Triop now has a smaller but still informative same-budget learned-baseline audit.
The compiled zero-retraining reference reaches \routeacc\ \(0.7979\), while plain LoRA only reaches \(0.3336\pm0.0006\), effectively chance.
Prompt-slot tuning improves substantially but remains below compiled at all retained slots: \(0.6818\pm0.0122\) at \texttt{a}, \(0.7089\pm0.0303\) at \texttt{b}, and \(0.6924\pm0.0084\) at \texttt{eq}.
Only LoRA plus trainable control embeddings matches the compiled row, reaching \(0.8036\pm0.0094\) under the same 800-step budget.
This makes triop a second routing branch in which the no-retrain versus learned-control distinction survives a matched learned-baseline audit, even though add/sub remains the broader relocation study.

\subsubsection{Qwen Local Routing}
Qwen local routing now has a small retained learned-relocation audit as well.
Within the audited three-site local family, the canonical compiled family \texttt{prompt\_last} remains sharply selected, with held-out rescue \(0.8603\), while the other compiled local families stay at \(0.0\).
Trainable local relocation behaves differently: all three learned sites reach held-out route accuracy \(1.0\) across the retained five-run sweep, and the best learned site \texttt{policy\_marker\_start} reaches learned delta \(6.4701\).
In the cost accounting used throughout the paper, the local learned rerun reaches the compiled score band after \(64\times100=6400\) support\(\times\)steps.
So the Qwen local branch matches the GPT-2 routing lesson qualitatively: the compiled family is sharply selected inside the audited local family, while trainable local relocation succeeds broadly once optimization is allowed.

\FloatBarrier
\subsection{Qwen Routing, Verify, and Solve}
This section collects the main Qwen numbers in one place so that the routing result and the reasoning-side limitations can be read without jumping across the main text.

\subsubsection{Routing}
Within the local routing family, the correct operator-token interface reaches perfect matched-interface routing, \(1.0\), while the wrong-interface prompt-last control falls to \(0.383\).
At layer 0, the \(QKV\) split keeps the same interpretation as the main text: \(V\)-only replacement closes at \(1.0\), \(K\)-only stays at \(0.0\), and matched random remains low at \(0.188\); receiver \(0.0\rightarrow1.0\) and donor \(1.0\rightarrow0.0\) swaps show that the state is transportable and behaviorally invertible at the tested site.
The cautionary numbers are also retained here: in the receiver-patched rerun, matched random \(KV\) rises to \(0.863\), and in the harder remap series it only falls gradually from \(0.852\) to \(0.664\), \(0.520\), and \(0.488\) while aligned \(V\)-only transfer remains perfect.
This is why the branch is reported as a strong local consistency check rather than a clean identity-level replication.

\subsubsection{Verify}
On verify, state-compiled no-CoT\(+\delta\) yields \(\Delta_{\texttt{patch-no}}=+0.1530\) and \(\Delta_{\texttt{patch-cot}}=+0.1847\), with positive deltas in all \(82/82\) evaluation runs.
The retained mediation boundary is weaker than routing closure: matched-random ablation drops only \(0.043\) with \(p=0.045\), and the best gain ratio is \(0.8667\) with \(k^*=\varnothing\).
So the appendix keeps the same branch-local conclusion as the main text: there is a real state-level effect under strict holdout, but not a near-sufficient matched-interface closure result.

\subsubsection{Solve}
On solve, the oracle ceiling remains large and stable: one-slot gains are \(+0.2267/+0.2333\) on two cohorts, while two-slot gains rise to \(+0.3933/+0.3767\).
The identifiability gap is the important retained point.
Top1 stays near \(+0.003/-0.033\), oracle@4 captures \(0.882\) of full gain, and learned commit reaches only \(0.183\) of oracle@4 while remaining close to oracle@1 at \(0.944\).
So shortlist quality is high, but committed control still behaves more like weak single-proposal selection than effective reranking.
The appendix retains this branch mainly to make the proposal-versus-commit separation numerically explicit.

\FloatBarrier
\subsection{Frozen-Grammar Cross-Task Audit}
This appendix block records the cleanest retained intermediate-complexity extension beyond the routing family.
The same \texttt{nodynamic perrow} router grammar is frozen and reused across held-out benchmark families rather than retuned task by task.

\subsubsection{Fixed-Grammar Cross-Task Readout}
On the retained fixed-grammar audit, compiled residual transport stays above the conflict receiver on both audited tasks, but the recovered fraction of donor routing is modest.
For \texttt{ifeval}, donor route accuracy is \(0.4902\), receiver is \(0.0117\), and compiled reaches \(0.0273\), giving gain \(0.0156\) and recovery fraction \(0.0327\).
For \texttt{beavertails}, donor is \(0.7910\), receiver is \(0.0\), and compiled reaches \(0.0820\), giving gain \(0.0820\) and recovery fraction \(0.1037\).
The mean recovery fraction across the retained tasks is therefore \(0.0682\).

\subsubsection{Interpretation}
This cross-task audit sits between the tight routing branches and the broader generation or solve settings.
The protocol remains informative beyond the original routing family, since compiled gains stay positive under the same frozen grammar, but the exact routing-style closure does not survive.
The retained lesson is that protocol portability extends farther than identity-level closure: the same intervention grammar still captures some useful state across tasks, but only a small fraction of donor routing behavior is recovered once task family changes.

\end{document}